\documentclass[journal]{journal}
\ifCLASSINFOpdf
  \usepackage[pdftex]{graphicx}
\else
\fi
\begin{document}
%
\title{Unsupervised Neural Architecture for Saliency Detection: Extended Version}
%
%
%

\author{Natalia~Efremova
        and~Sergey~Tarasenko
\thanks{N. Efremova is with Plekhabov's University, Moscow, Russia.
E-mail: natalia.efremova@gmail.com.}
\thanks{S. Tarasenko is independent researcher.  Email: infra.core@gmail.com
}
}

\maketitle
\thispagestyle{empty}

\begin{abstract}
We propose a novel neural network architecture for visual saliency detections, which utilizes neurophysiologically plausible mechanisms for extraction of salient regions. The model has been significantly inspired by recent findings from neurophysiology and aimed to simulate the bottom-up processes of human selective attention. Two types of features were analyzed: color and direction of maximum variance. The mechanism we employ for processing those features is PCA, implemented by means of normalized Hebbian learning and the waves of spikes. To evaluate performance of our model we have conducted psychological experiment. Comparison of simulation results with those of experiment indicates good performance of our model.
\end{abstract}

\begin{IEEEkeywords}
neural network models, visual saliency detection, normalized Hebbian learning, Oja's rule, psychological experiment
\end{IEEEkeywords}

%
\IEEEpeerreviewmaketitle

\section{Introduction}
%
%
%
%

\IEEEPARstart{P}{recise} processing of only important regions of visual scenes is one of essential properties of human visual system. Extraction of important regions is referred to as \textit{selective attention}: ``The mechanism in the brain that determines which part of the multitude of sensory data is currently of most interest is called selective attention" \cite{Frintrop_survey} (p. 6).

To date, two types of attention have been distinguished: bottom-up and top-down. Bottom-up attention is believed to be triggered only by characteristics of the visual scene. Top-down attention is hypothesized to be driven by higher cognitive processes (individual's previous experience, expectations and current goals).

In this study, we only concentrate on bottom-up selective attention, which is also referred to as $saliency$. Frintrop et al. \cite{Frintrop_survey} (p. 6) defines saliency as follows: ``Regions of interest that attract our attention in a bottom-up way are called $salient$ and the responsible feature for this reaction must be sufficiently discriminative with respect to surrounding features".

In this paper, we introduce an unsupervised neural architecture, which detects the salient regions in visual scenes in neurolophysiologically plausible way on the one hand, and in the computationally inexpensive manner on the other hand. To achieve these goals, we use three main neurophisiological findings which comprise a neurolophysiological plausibility of our model: 1) findings by Knierim and Van Essen \cite{kness}; 2) waves of spikes \cite{thorpe1,thorpe2,thorpe3}; and 3) normalized Hebbian learning \cite{hebb,oja82,rummc}.

It was illustrated in the seminal work by Knierim and Van Essen \cite{kness} that activation of neurons in primary visual area V1 of macaque monkey depends not only on the actual stimulus, presented in the receptive field (RF) of a given neuron (stimulus 1), but also on how different are the stimuli presented to neighbour neurons from the stumilus 1. Knierim and Van Essen found that a neuron produced lower response if the stimulus is the same within the stimuli of the surrounding neurons comparing to the case than stimuli of the surrounding neurons are different. Therefore neuron with a stimulus outstanding from the surrounding texture will procduce higher response. The work by Knierim and Van Essen \cite{kness} is considered to provide insight into $saliency$ mechanism on the neural level \cite{li}.

Concept of \textit{waves of spikes} has been introduce by Van Rullen and Thorpe \cite{thorpe3}. The core of this concept is that the stronger response is produced by a neuron faster than the weaker one. Therefore neurons fire with different speed. The responses of conherently firing neurons form waves of spikes. 

Mechanism of attention based on waves of spikes has been proposed by Van Rullen and Thorpe \cite{thorpe3}. They described the rapid and reliable mechanism  of processing natural input scenes by ventral stream of the visual system, based on the relation between stimulus saliency and spike relative timing. According to their theory, waves of spikes are generated in the retina in response to a visual stimulation and the obtained information is transferred further explicitly in its spatio-temporal structure: the most salient information is represented by the first spikes over the population. 

The next mechanism we utilize is principal component analysis (PCA) that was proven to be effective instrument for saliency detection \cite{zhouetal, yuetal}. However, we suggest more neurophysiologically plausible mechanism for implementation of PCA: we employ Oja's rule to extract the first principle component in a neurophysiologically plausible way \cite{oja82}.

The rest of the study is organised as following: first, we describe the architecture of the proposed model and the implementation of the above mechanisms in our model. Next, we introduce the psychological experiment with human subjects and comparison of experimental results with results produced by our model. Finally, we evaluate the performance of our model and discuss the future work.

\section{Model Description}
In this section, we describe in details each computational principle we employ, and the way how all three principles are integrated together to deliver simple neurophysiologically plausible mechanism of saliency detection. 

\subsection{Unfold Static Image in Time by using Waves of Spikes concept}

If we consider ``integrate-and-fire" model of a neuron, then we can expect that time at which a neuron reaches its threshold is a monotonous (decreasing) function of its activation (i.e. the more a neuron is activated, the sooner it fires). The logic behind this assumption is as follows. A neuron integrates its inputs over time until it reaches a threshold, and fires a single spike (action potential). After a certain refractory period, neurons starts functioning again. This property has been usedto support the idea that the firing rate of a neuron is a monotonous function of the strength of its input (i.e. the more a neuron is activated, the more it fires). From the point of time domain, the time at which a neuron reaches its threshold is also a monotonous (decreasing) function of its activation (i.e. the more a neuron is activated, the sooner it fires). This means that the latency of firing of a neuron, just as much as its firing rate, will reflect the strength of its input.

Therefore if stimuli are presented to the population of neruons, the first spike of the population corresponds to the most activated neuron, the second spike to the next most activated neuron, etc.  This idea is the basis of the \textit{Rank Order Coding Scheme} \cite{thorpe1, thorpegautrais97, thorpegautrais98}. 

The neurons which fire first could be considered as representing the highest intensity values of the stimulus, it is better to think of them as carrying the \textit{most salient information}. In the retina for example, the stimulus property that determines the activation level, and thus the latency of firing of a neuron, is not luminance per se, but rather luminance contrast. It is well known to experimenters in psychophysics and electrophysiology \cite{reynoldsetal} that the primate visual system interprets stimulus contrast as a primarydeterm inant of stimulus saliency, and there is plentyof data showing that latency varies with stimulus contrast \cite{gawneetal}.

The neurons, whose actual visual input is closer to the preferred stimulus, fire faster \cite{thorpe3}. Therefore, the higher is the activation of neuron the faster it fires. In \cite{tarasenko}(p. 1518), the following approach for modelling such process by means of non-spiking neural networks was suggested: ``...The neurons with unit activation (response is 1 on the normalized scale) fire first. Then the neurons with activation greater or equal to 1-$\epsilon$ ($\epsilon$ = 0.1), excluding previously fired neurons, will fire and so on. Thus, the neurons with the same activation level form `waves of spikes'. Consequently, the original retinal image is unfolded in a time domain."

In the case of color intensity, we consider that the signals corresponding to the higher color intensities are produced faster than signal corresponding to the lower color intensities. First the highest color intensity (unit amplitude on the normalized scale) signal will be generated. 

To employ the concept of \textit{waves of spikes} in this study, we consider color RGB image as three independent R-, G-, and B-channels. Then we decompose signal of each channel into 10 intensity levels starting from 0 with step $\epsilon$. Thus, the first level contains color intensities from the interval [0, 0.1), second level contains intensities from the interval [0.1, 0.2), etc.

\begin{figure}
\centering
\includegraphics[width=9cm]{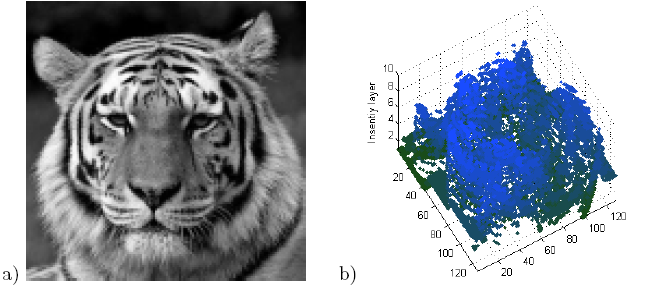}
\caption{Black and white image decomposition into intensity layers: a) original black and white; b) 3d decomposition using intensity layers.}
\label{tiger}
\end{figure}

\begin{figure}
\centering
\includegraphics[width=9cm]{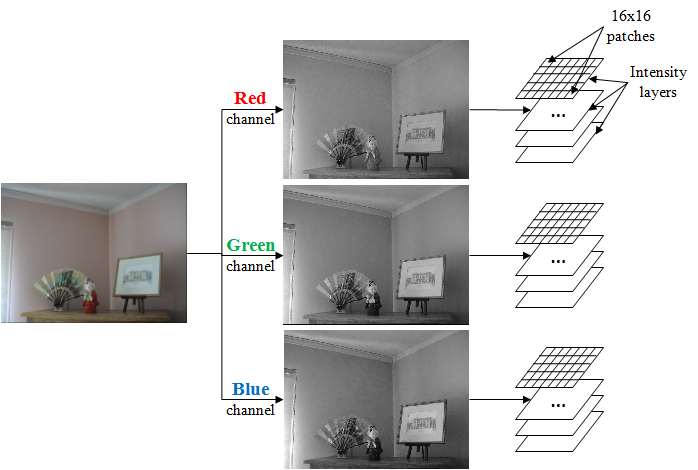}
\caption{Image pre-processing pipeline. First three difference color channel (i.e., Red, Green and Blue) are considered. Then intensity levels are applied to each color channel.}
\label{process}
\end{figure}

Therefore it is possible to decompose entire image into several layers (levels of intensities). This is illustrated in Fig.~\ref{tiger}. In out study we use color images. For color images three separate color channels are used. Then intenstity levels are applied to each of three color channels. Entire pipeline for image pre-processing in our study is presented in  Fig.~\ref{process}.

After such decomposition, we split each intensity layer for each color channel into 16 x 16 pixels patches. Each 16 x 16 patch is considered to be a RF of a visual neuron, which extacts the first (main) principle component by means of neurophysiologically plausible analogue of PCA. In the next section, we consider such neurophisiologically plausible analogue in greater details.

\begin{figure}
\centering
\includegraphics[height=5cm]{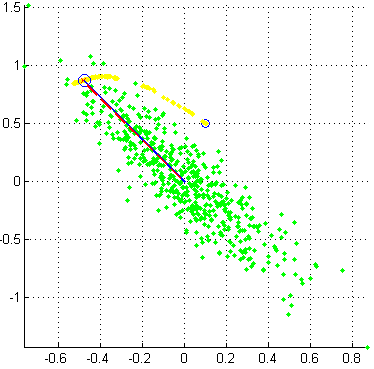}
\caption{Illustration of the first principle component generate by PCA and Oja's rule for 500 random 2D data samples. Yellow dots illustrate convergence trajectory of originally set weight vector $w=[0.1000, 0.5000]$ to weight vector  $w^* = [-0.4805,0.8773]$ after Oja's learning. $w$ converges to $w^*$ after 2500 iterations (each sample is presented five times). True $1^{st}$ principle component $pc_1 = [-0.4847,0.8747]$. Angle between $w^*$ and $pc_1$ is $0.2830^{\circ}$. Length of $w^*$ is 1.0024. Therefore weight vector $w^*$ is almost perfect approximation for the $1^{st}$ principle component $pc_1$. In this case, inputs into Oja's rule are $x$- and $y$-coodinates of the 2D points. The $1^{st}$ and $2^{nd}$ components of weight vectors $w$ and $w^*$ are $x$- and $y$-coodinates, respectively. To graphically illustrate $1^{st}$ principle component $pc_1$ and its approximation $w*$, two radius-vectors with ending points having coordinates of $pc_1$ and $w^*$, respectively, are used.}
\label{pcavsOja}
\end{figure}

\subsection{PCA for saliency and neural implementation of PCA: Hebbian learning and Oja's rule}

PCA was used by Zhou et al. \cite{zhouetal} in conjunction with distance computation and histogramming methods to detect salient regions. The goal of our study is to provide neurophysiologically plausible algorithm (PCA itself is not plausible enough for our purposes). On the other hand, Hebbian learning \cite{hebb} and normalized Hebbian learning (Oja's rule) \cite{oja82} are biologically plausible analogues of PCA. Hebbian learning and Oja's rule deliver the first principle component into weights of a given neuron tuned by such learning rule.

\textit{Hebbian learning}. Hebb has suggested that certain cortical cell populations behave as functional units with coordinated activity patterns in accordance with changing synaptic strength \cite{hebb}. It was assumed that the strength of a synapse (connection) between \textit{neuron} \textit{1} and \textit{neuron} \textit{2} increases, when firing in neuron 1 is followed by firing in neuron 2 with a very small time delay. 

The discrete Hebbian learning rule for a single neuron can be presented in the form of eq. (\cite{hebb}):
\begin{equation}
\Delta w(k+1) = \mu [y(k)x(k) - \alpha w(k)] 
\label{eq_hebb1}
\end{equation}
where $y$ is output of the neuron, $w=(w_1,...,w_n)$ is vector of synaptic weights, $\mu$ and $\alpha$ are constants regulating learning and forgetting rates, respectively, and $k$ indicates the iteration number.

The output $y$ can  be calculated as $y = w^Tx$, where $w=(w_1,...,w_n)$ is a vector of synaptic weights, $x=(x_1,...,x_n)$ is vector of inputs, and $n$ is a total number of inputs of the given neuron. 

The equation for calculation of the synaptic weights update can be formulated as following  
\begin{equation}
w(k+1) = w(k) + \mu(y(k)x(k) - \alpha w(k)) 
\label{eq_hebb3}
\end{equation}

\textit{Oja's learning rule}. The main shortcoming of this learning rule (eq. \ref{eq_hebb3}) is unlimited growth of synaptic weights. To overcome this problem, Oja \cite{oja82} has proposed normalized Hebbian or Oja's rule:
\begin{equation}
w(k+1) = w(k) + \mu (y(k)x(k) - y(k)^2 w(k) )
\label{eq_oja}
\end{equation}

The final weight vector after learning will represent the first (main) principle component of the presented data. In this study, we employ Oja's rule to evalute main principle component of the color intensity presented in each patch.

\begin{figure}
\centering
\includegraphics[height=5cm]{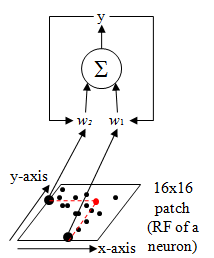}
\caption{Stage 1: Oja's learning. Use $x$- an $y$- coordinates of the non-zeros intensity pixels to compute approximation of the first principle component for the non-zero intensity pixels within the given 16x16 patch. The examplar pixel is marked as a red point. During this stage output $y$ of a neuron is used to update synaptic weights $w_1$ and $w_2$.}
\label{stage1}
\end{figure}

\begin{figure}
\centering
\includegraphics[width=8cm]{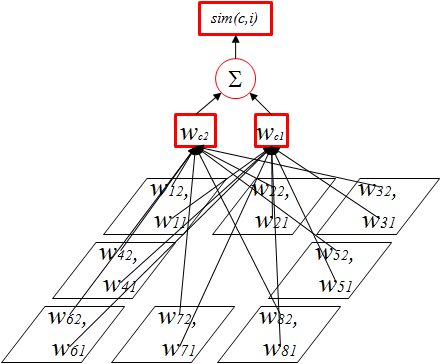}
\caption{Stage 2: Lateral interaction for similarity computation. Similarity is considered as a dot product between the weight vector of the center neurons and weights of the surrounding neurons. For each pair of the center neuron and one of the surrounding neurons similarity measure is computed separately.}
\label{stage2}
\end{figure}

\subsection{Identifying salient regions}
In the work by Knierim and Van Essen \cite{kness}, it was shown that neural responses depend on the activation of surrounding neurons. Therefore it is necessary to compare how similar the stimulus the neuron at the center is to the stimuli of the surrunding neurons. 

Hebbian learning principle was  employed for saliency detection Yu et el. \cite{yuetal}. They use spatial correlation measure to compare different regions of an image. 

In our study, we consider that the contents different 16x16 pixel patches can be represented by their fisrt principle components. Then the more similar the contents are the higher the value of the dot product of two first principle components.


This approach is implemented in two stages. Stage 1 is tuning the weight vector of each neuron to approximate the first principle component of the a 16x16 pixel patch (RF of a neuron). Since all synaptic weight vectors are the first principle components (orthonormal vectors), therefore the output of the a neuron is cosine of an angle between priciple component and the input. If the input is a synaptic weight of another neuron, then the output of the give neuron is is cosine of an angle between two priciples components of two neurons, i.e., two patches. We consider the cosine of such angle to characterize the degree of similarity between two patches. 

In other words, by measuring the cosine of angle between two synaptic weights (the first priciple components) we evalute how similar are original stimuli at the RFs of the two given neurons. Therefore in Stage 2 we compare similarity of stimili of surrunding neurons by means of computing dot product between the weight vectors of the center neuron and the surrouding ones:
\begin{equation}
sim(c,i) = w_c^Tw_i 
\label{sim}
\end{equation}
where $w_c$ and $w_i$, $i=1,...,8$, are weight vectors of the center neuron and the surrounding neurons after stage 1. This similarity measure is computed for each pair of the center neurons with one of the surrounding neurons.


Summarizing, during Stage 1 a neuron takes its input from a RF (16x16 patch) for the purpose of Oja's learning.  At this stage the inputs into the neurons are $x$- and $y$-coordinates of pixels with non-zero intensities in the 16x16 pixel patch at a certain intensity layer. All the pixels within the patch at a particular intensity layer are considered to be of the same intensity. Therefore a single 16x16 pixel patch is considered to be 16x16 binary matrix with 1 and 0 meaning non-zero and zero intensities, respectively.

After learning is completed, during Stage 2,  this neurons takes inputs from all surrounding neurons with stimuli being synaptic weights (approximations of the first principle components) of surrounding neurons. In this case, there are eight possible connections between the given neuron at the center (the center neuron) and surrounding neurons.

By computing similarity with surrounding neurons, we estimate how homogeneous the vicinity of the given center neuron. We set the decision rule to distinguish between similar and dissimilar stimuli:  if $ w_c^Tw_i < 0.1$, then we consider that contents in the RFs of the center neurons and $i$-th neuron are dissimilar.

To process a stimulus in the RF of a given neuron, we compute first principle component for each color channel for the given patch. Then we calculate the total number of dissimilar weight vectors in a surrounding neurons for each color channel separately. If the number of dissimilar weight vectors across all intensity level for a certain color channel is greater than a threshold, we consider that given stimulus is a \textit{salient region} for the given color channel. In this study, we use threshold value equal to 10.

Finally, throughout the color channels we calculate frequency of a certain patch to contain salient regions. At this point we cut off the patches with frequencies less than an expected value by chance. The expected value is calculated as a ratio of total number of patches containing salient region across color channels and intensity layers to the total number of patches to cover the figure.

\section{Psychological Experiment}

\subsection{Method}
\textit{Objective}. In this study, we aim to show that our model can predict the location of salient regions in natural scenes and home scenes in the similar way human subjects do. To evaluate performance of our model, we have conducted a psychological experiment to collect the data about performance of human subjects.

\textit{Participants}. All participants were undergraduate students (ages 18$-$23) at the Plekhanov's University, who received course credit for participation. All participants had normal or corrected-to-normal vision. In total 18 subjects were tested, including 11 males and 7 females. 

\textit{Procedure}. During the experiments, we asked participants to select all the regions, which attracts their attention on the given set of images. The images are taken from the database, provided by Bruce and Tsotos \cite{bruce2006,bruce2009}. In particular, we used 21 images numbered 20 to 40 in Bruce original enumeration. In this study we enumerate images starting from 1 to 21 beginning with image 20.

All the subjects were given the instructions to select all the regions of the image that attract their attention. They were ask to do it by using mouse to encompass the Region Of Interest (ROI) with a color curve on computer screen. Each ROI was defined as an area within the mouse selection. 

\subsection{Results of Psychological Experiment}
For the purpose of further analysis, we have discarded results of two subjects. One subject has not selected any region in either of images. The regions, selected by another subject, were too large for our purposes (over 30 percent of a visual scene). We have also discarded some results of other participants for the same reason.

Finally, we have obtained from 12 to 16 individual ROI maps for each image. To obtain final integrated ROI map we have superimposed individual ROI maps for a single image produced by all subjects. Therefore integrated ROI map illustrates frequencies of selection of specified ROIs in a visual scene.

\begin{figure}
\centering
\includegraphics[width=5cm]{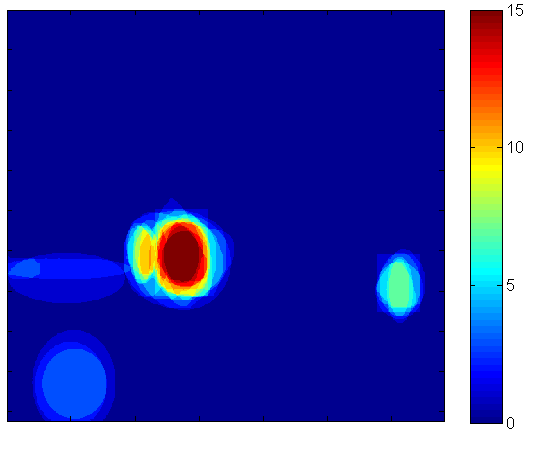}
\caption{The results of the psychological experiment for one of the images}
\label{fig:result_of_experiment}
\end{figure}

The sample integrated ROI map for one of the images is presented in Fig. \ref{fig:result_of_experiment}. It demonstrates the selected regions for all of the tested subjects (heat map), where red color represents the most  frequenctly selected region of the scene and the blue color corresponds to the least frequently selected region of the scene. This integrated ROI map is for image 32 in Bruce and Tsotos \cite{bruce2006,bruce2009} original enumeration.

\section{Evaluation of the proposed model}
\begin{figure}
\centering
\includegraphics[width=9cm]{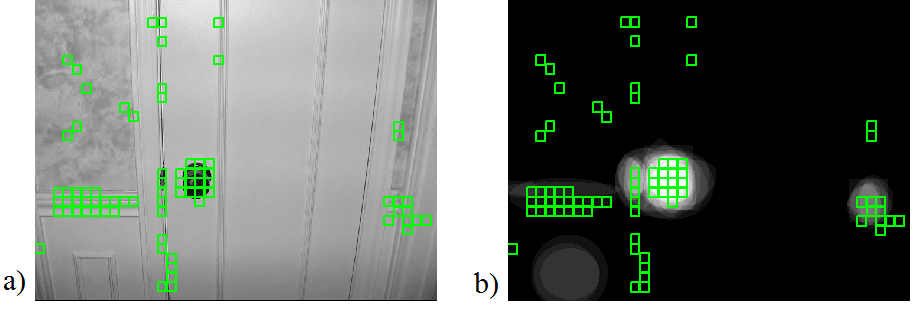}
\caption{The results of the psychological experiment for one of the images compared to the saliency regions, detected by the proposed model. a) Simulation results are superimposed on the original image; b) the simulation results are superimposed on the behavioural data.}
\label{fig:result_comparison}
\end{figure}

To evaluate performance of our proposed model, simulation results produced by our model were compared with those of the human subjects. On Fig. \ref{fig:result_comparison} the simulation results are compared with the results of the psychological experiment for image 32.  Complete results of model performance for each image vs integrated ROI map is presented in Table \ref{table2}.

\begin{table}[hc]
\caption{Evaluation of model performance}
\centering
\begin{tabular}{c|c|c|c|c} \hline
Image  & Recall & Precision & Weighted & Weighted \\
Number &  & & Recall & Precision \\ \hline
Image 1 & 0.925 & 0.315 & 0.944 & 0.375 \\
Image 2 & 0.914 & 0.275 & 0.938 & 0.486 \\
Image 3 & 0.797 & 0.431 & 0.827 & 0.398 \\
Image 4 & 0.956 & 0.302 & 0.967 & 0.511 \\
Image 5 & 0.728 & 0.399 & 0.774 & 0.567 \\
Image 6 & 0.631 & 0.162 & 0.701 & 0.284 \\
Image 7 & 0.416 & 0.448 & 0.486 & 0.425 \\
Image 8 & 0.330 & 0.434 & 0.360 & 0.562 \\
Image 9 & 0.369 & 0.489 & 0.419 & 0.404 \\
Image 10 & 0.505 & 0.426 & 0.537 & 0.460 \\
Image 11 & 0.658 & 0.218 & 0.734 & 0.411 \\
Image 12 & 0.973 & 0.185 & 0.982 & 0.472 \\
Image 13 & 0.533 & 0.143 & 0.617 & 0.339 \\
Image 14 & 0.865 & 0.450 & 0.887 & 0.599 \\
Image 15 & 0.868 & 0.423 & 0.887 & 0.478 \\
Image 16 & 0.872 & 0.147 & 0.898 & 0.273 \\
Image 17 & 0.953 & 0.169 & 0.970 & 0.451 \\
Image 18 & 0.926 & 0.173 & 0.949 & 0.273 \\
Image 19 & 0.743 & 0.520 & 0.764 & 0.556 \\
Image 20 & 0.849 & 0.371 & 0.864 & 0.379 \\
Image 21 & 0.686 & 0.383 & 0.715 & 0.508  \\ \hline
Average & 0.734 & 0.327 & 0.772 & 0.439  \\ \hline
\end{tabular}
\label{table1}
\end{table}

To evaluate the performance of our model, we compared the salient regions selected by our model with ROI maps. To measure performance, we calculate precision and recall rates. Precision is the fraction of retrieved instances that are relevant, while recall is the fraction of retrieved relevant instances \cite{Rijsbergen}. 

\begin{table}[hc]
\caption{ROI maps and model performance}
\begin{tabular}{cc} 
\includegraphics[width=4cm]{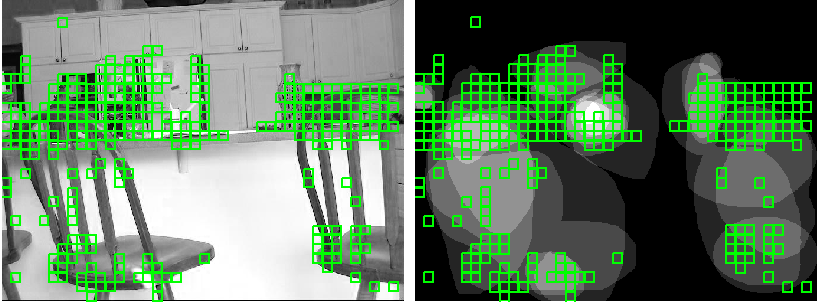} & \includegraphics[width=4cm]{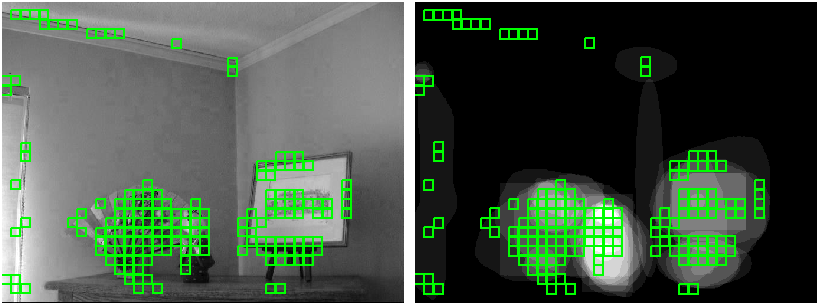} \\ 
\includegraphics[width=4cm]{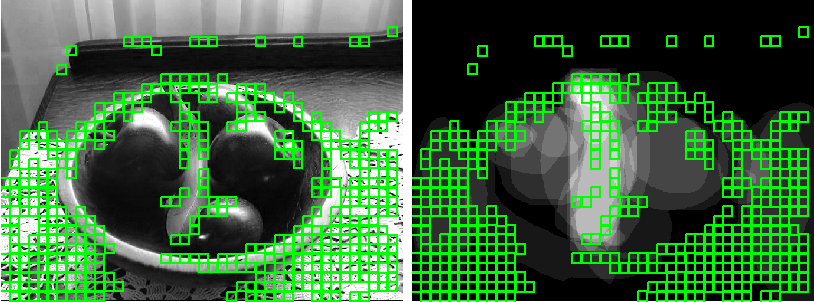} & \includegraphics[width=4cm]{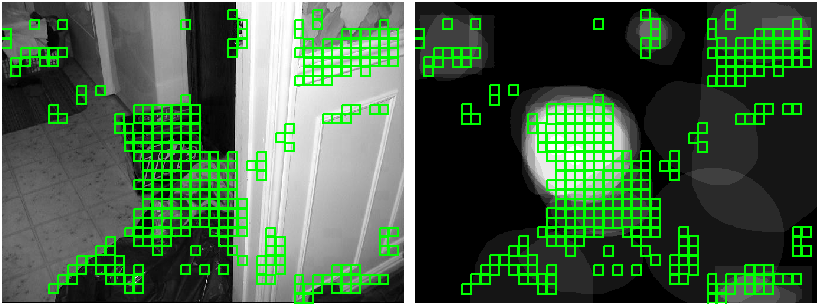} \\ 
\includegraphics[width=4cm]{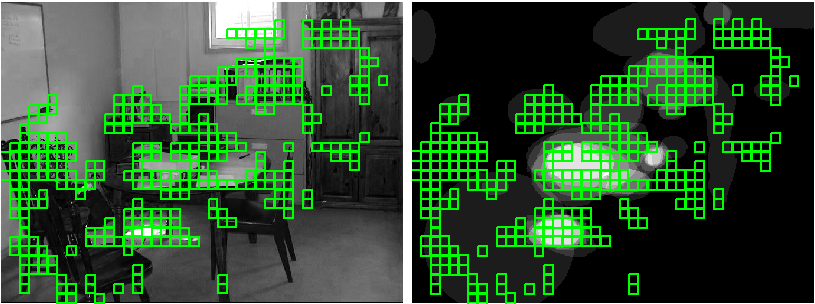} & \includegraphics[width=4cm]{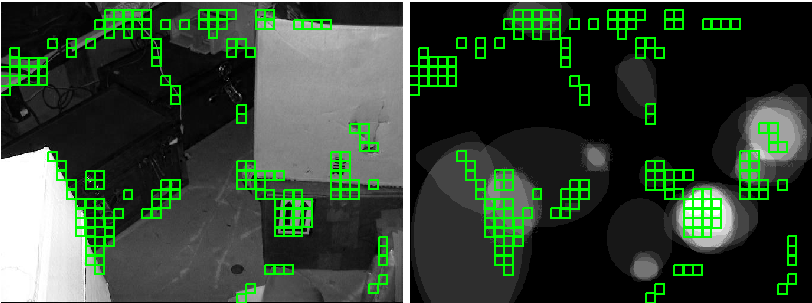} \\
\includegraphics[width=4cm]{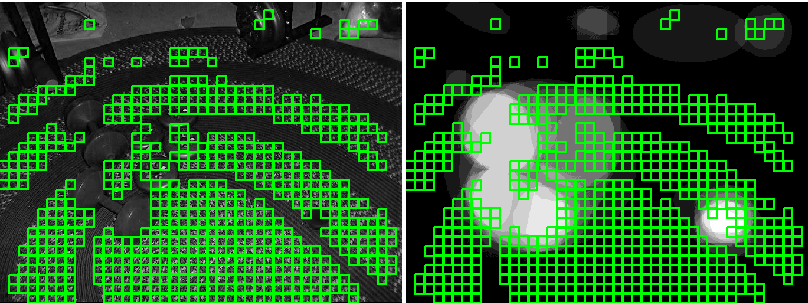} & \includegraphics[width=4cm]{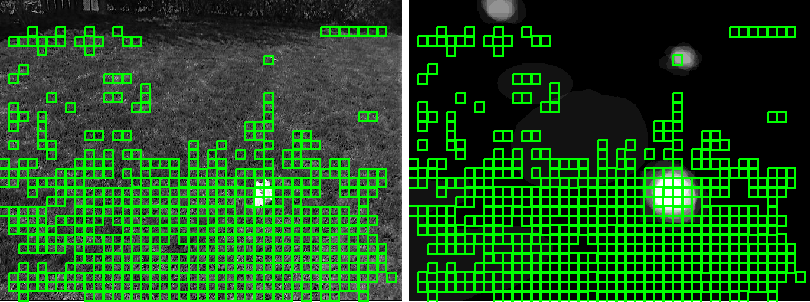} \\
\includegraphics[width=4cm]{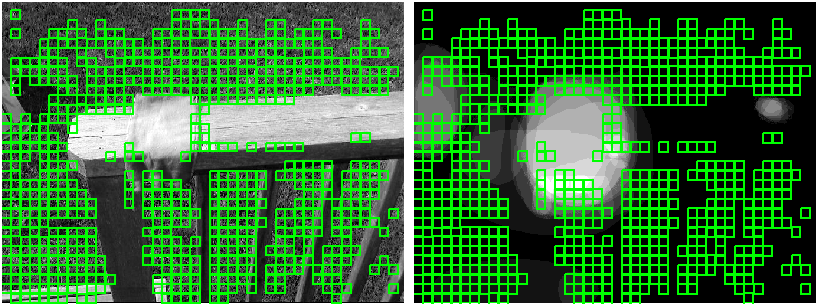} & \includegraphics[width=4cm]{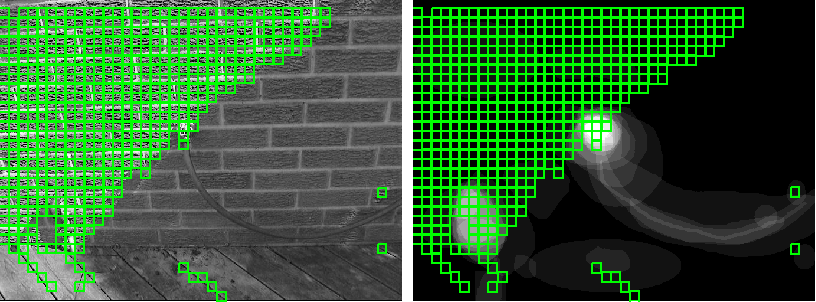} \\
\includegraphics[width=4cm]{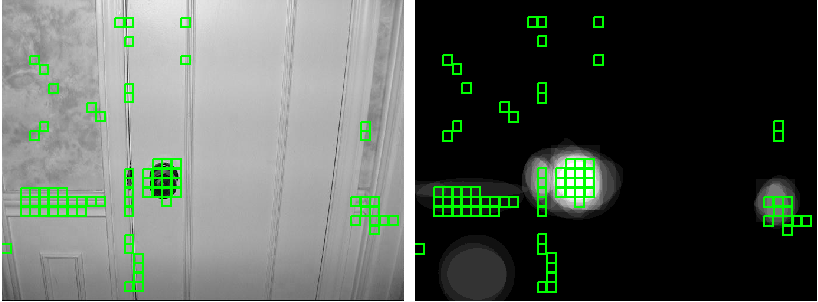} & \includegraphics[width=4cm]{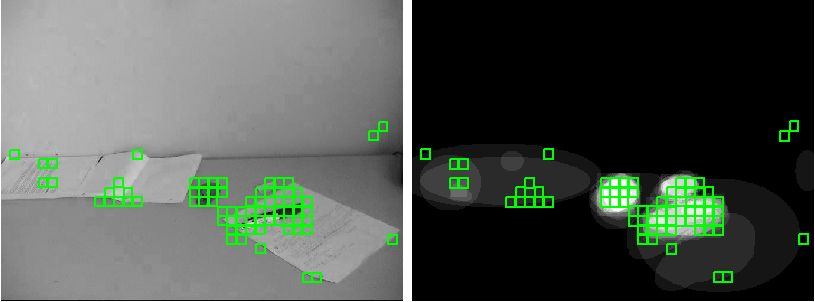} \\
\includegraphics[width=4cm]{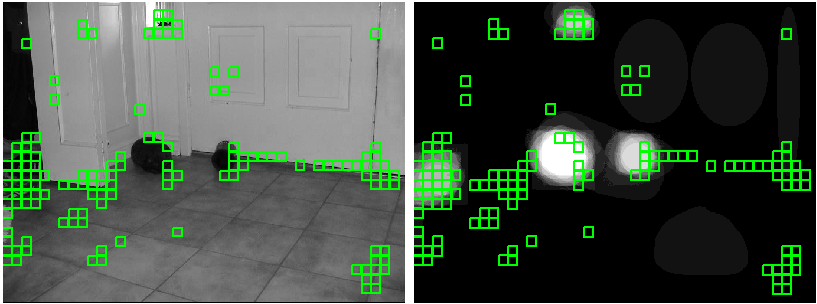} & \includegraphics[width=4cm]{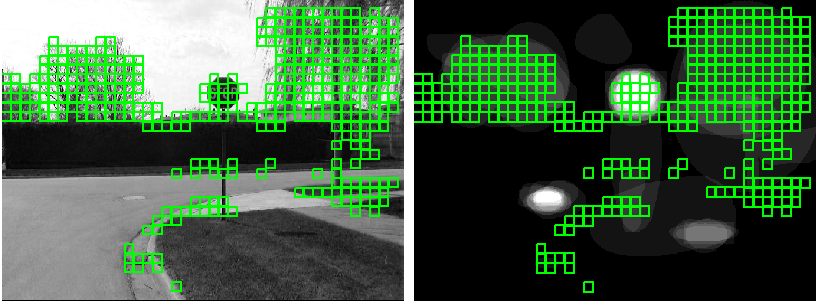} \\
\includegraphics[width=4cm]{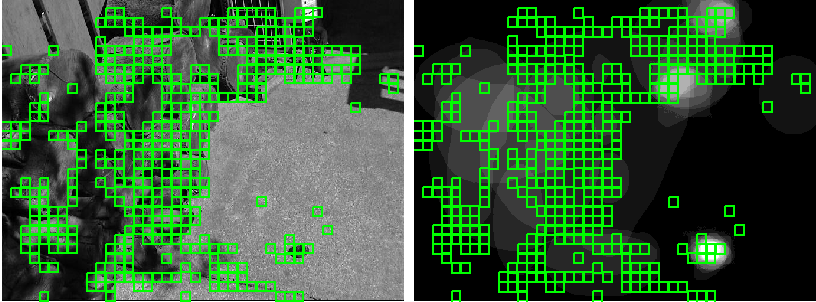} & \includegraphics[width=4cm]{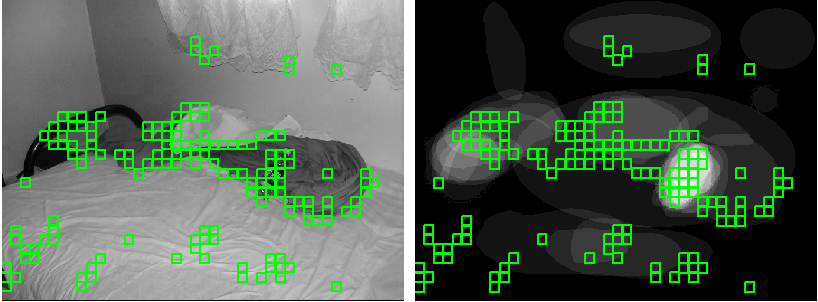} \\ 
\includegraphics[width=4cm]{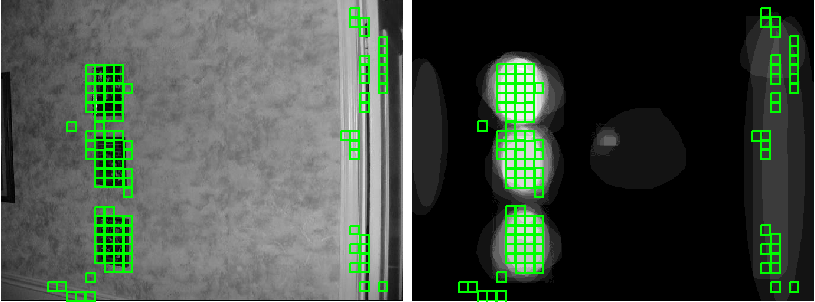} & \includegraphics[width=4cm]{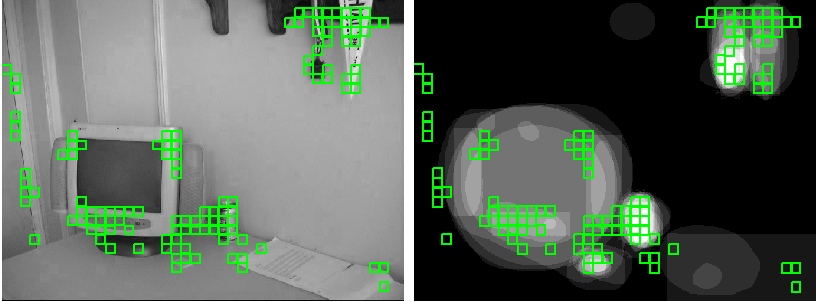} \\
\includegraphics[width=4cm]{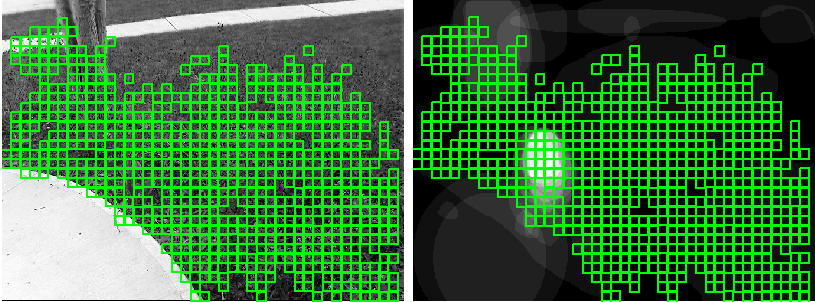} & \includegraphics[width=4cm]{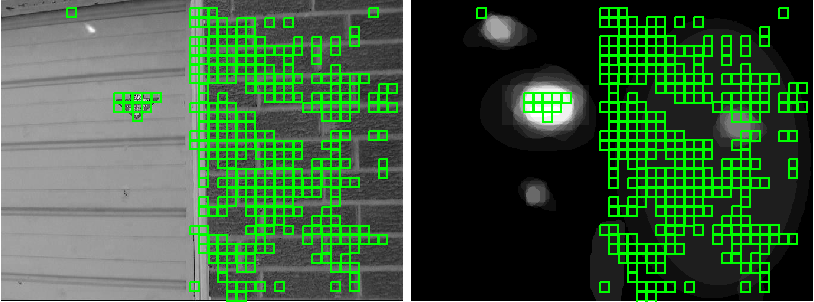} \\
\includegraphics[width=4cm]{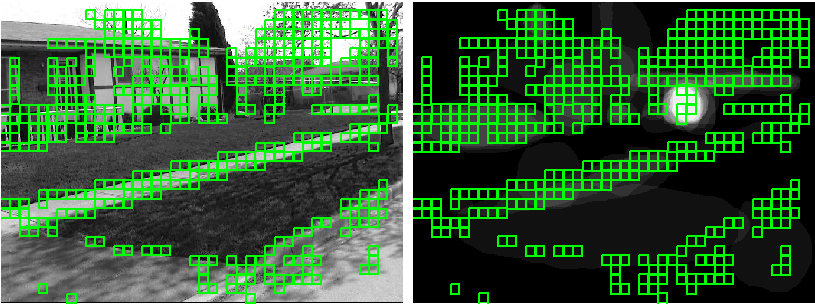} & \\
\end{tabular}
\label{table2}
\end{table}

The total number of patches, which are selected by our model, and contain at least one non-zero value of integrated ROI map, we consider as \textit{true positives (tp)}. While the patches selected by our model, but containing only zero values, we consider as  \textit{false positive (fp)}. The number of patches, which contain at least one non-zero value, and were not selected by our model, we consider to be \textit{false negative (fn)}.

Then precision is calculated as $precision$ = $\frac{tp}{tp+fp}$. 

Recall is calculates as $recall$ = $\frac{tp}{tp+fn}$.

The integrated ROI maps are frequency maps that illustrate how many participants have selected particular regions.
Therefore the regions, selected by the majority of subjects (with high frequencies) can be considered more important than others. Therefore, detection of more frequently selected regions demonstrate better model performance than detection of less frequent ones. 

To take this idea of region importance due to selection frequency into account, we also calculate weighted precision and weighted recall along with standard ones. By normalizing the integrated ROI maps, we obtain probability density of region selections. Therefore, the sum of all non-zero values should equal to 1. This probability distribution represents all positive samples. 

In this context, $precision$ is a part of probability density contained in patches, which were selected as salient ones by our model. We refer to $precision$ defined in this way as \textit{weighted precision}.

However, calculation of recall is not that straight forward, because all false positive patches contain zero values. To calculate \textit{weighted recall}, we first add 1 to all the non-normalized integrated ROI map values. Then we calculate the sum of all integrated ROI elements in $tp$ and $fp$ patches.

The results of calculation of $recall$, $precision$, \textit{weighted recall} and \textit{weighted precision} are presented in Table \ref{table1}. 

For all images, the value of weighted recall is higher that of standard recall. The weighted precision values is higher than standard precision values for all images except for images 3, 7 and 9. The mean recall and precision values are 0.734 and 0.327, respectively. The mean weighted recall and weighted precision values are 0.772 and 0.439. This results indicate that on average our model selects more frequent salient regions.

Overall results suggest that our model was quite accurate at predicting human judgements of
ROIs in the visual scenes. 

\section{Discussion and Conclusion}
In this study, we have proposed a neurophisiologically plausible neural architecture for saliency detection. The main specifics of our model is that it is contracted in a completely unsupervised manner. The computational principles, which were utilized for building this model, are considered to be employed by the human brain. Finally, our model demonstrates good performance of saliency detection based on only local information and is less computationally greedy then other proposed algorithms.

The observation of model performance illustrates that the proposed model is especially good in detecting salient regions surrounded by homogeneous  areas. However, model produces a lot of false positive responses when dealing with high-contrast textures surfaces if they are extensively presented on the image (e.g. grass, carpets, structured bricks etc.). This restriction of a model is a subject for future work. 
\ifCLASSOPTIONcaptionsoff
  \newpage
\fi
\end{document}